\documentclass[11pt]{article}

\usepackage[]{acl}

\usepackage{times}
\usepackage{latexsym}
\usepackage{multicol}
\usepackage{multirow}
\usepackage{natbib}
\usepackage[T1]{fontenc}

\usepackage[utf8]{inputenc}
\usepackage{pgfplots}
\pgfplotsset{compat=1.18}
\usepackage{microtype}

\usepackage{inconsolata}

\usepackage{graphicx}

\title{Why Expert Alignment Is Hard: Evidence from Subjective Evaluation}

\author{
Tzu-Mi Lin$^{1}$, Wataru Hirota$^{2}$, Tatsuya Ishigaki$^{3}$, Lung-Hao Lee$^{1}$, Chung-Chi Chen$^{3,4}$ \\
$^{1}$National Yang Ming Chiao Tung University, Taiwan \\
$^{2}$Stockmark, Japan \\
$^{3}$National Institute of Advanced Industrial Science and Technology (AIST), Japan \\
$^{4}$National Institute of Informatics (NII), Japan \\
\texttt{ltmdegf4.ii12@nycu.edu.tw, wataru.hirota@stockmark.co.jp} \\
\texttt{ishigaki.tatsuya@aist.go.jp, lhlee@nycu.edu.tw, chen@nii.ac.jp}
}

\begin{document}
\maketitle
\begin{abstract}
Aligning large language models with expert judgment is especially difficult in subjective evaluation tasks, where experts may disagree, rely on tacit criteria, and change their judgments over time. In this paper, we study expert alignment as a way to understand this difficulty. Using expert evaluations and follow-up questionnaires, we examine how different forms of expert information affect alignment and what this reveals about subjective judgment.
Our findings show four consistent patterns. First, alignment difficulty varies substantially across experts, suggesting that expert evaluation styles differ widely in their distance from a model's prior behavior. Second, explicit criteria and reasoning do not always improve alignment, indicating that expert judgment is not fully captured by verbalized rules. Third, editing is sensitive to both the number and the identity of examples, with small numbers of edits providing useful but unstable gains. Fourth, alignment difficulty differs across evaluation dimensions: dimensions grounded more directly in proposal content are easier to align, while dimensions requiring external knowledge or value-based judgment remain harder.
Taken together, these results suggest that expert alignment is difficult not only because of model limitations, but also because subjective evaluation is inherently heterogeneous, partly tacit, dimension-dependent, and temporally unstable.
\end{abstract}

\section{Introduction}
\label{sec:introduction}

Human judgments are widely used to evaluate ideas, plans, and system outputs.
In many domains, expert opinions serve as the practical ``ground truth''.
Examples include proposal evaluation, peer review, product assessment, and policy analysis.
In such settings, decisions often depend on experience and professional judgment rather than strictly defined rules.
However, expert evaluation is rarely stable or uniform.
Different experts can assign very different scores to the same proposal.
Even the same expert may change their assessment over time.
These differences are especially common when the task involves subjective factors such as novelty, usefulness, or market potential.
As a result, expert evaluation is often heterogeneous and context-dependent.

With the rapid progress of large language models (LLMs), researchers have begun to explore whether models can act as automated evaluators.
LLMs are attractive for this purpose because they are fast, scalable, and consistent.
A common approach is to align models with expert judgments using expert-labeled data.
This can be done through prompting, in-context learning, fine-tuning, or preference alignment.
Recent work also studies personalized alignment, where models follow the preferences of a specific user or persona \cite{Chen2024,Zhang2025}.
Most alignment approaches implicitly assume that the target behavior is well-defined.
In other words, they assume that the expert evaluation rule can be learned from data.
For subjective tasks, this assumption may not hold.
Expert judgments may rely on tacit knowledge, contextual reasoning, or value trade-offs that are difficult to formalize.
This raises a fundamental question:
\emph{why is expert alignment difficult in subjective evaluation tasks?}

In this paper, we treat expert alignment not only as an engineering goal but also as a lens for studying subjective evaluation.
Our study is built on the PBIG 2025 shared task, where experts evaluate business proposals generated from patents \cite{Hirota2025}.
PBIG provides the base expert evaluation dataset across several dimensions, including technical validity, innovativeness, specificity, market size, and competitive advantage.
On top of this existing dataset, we augment the data with expert-level questionnaire annotations.
The questionnaire collects expert background information and asks experts to describe their evaluation criteria and reasoning processes.
This augmentation allows us to examine the gap between what experts \emph{do} (their scoring behavior) and what they \emph{say} (their explicit criteria and explanations).

We compare three families of alignment methods:
prompting (zero-shot and few-shot),
fine-tuning with parameter-efficient updates \cite{Hu2021},
and model editing \cite{Fang2024}.
For model editing, we treat an expert's evaluation style as knowledge to inject into the model.
Importantly, we vary the type of expert information used during editing:
(1) background information only,
(2) background with stated evaluation criteria,
and (3) background with criteria and example-based reasoning.
Our experiments reveal several consistent patterns that help explain why expert alignment is difficult in subjective settings.
First, experts vary widely in how close their evaluation behavior is to the base model.
Some experts require only small adjustments, while others require large behavioral shifts.
Second, we observe a strong ``method asymmetry.''
Model editing often produces small but reliable changes, whereas fine-tuning can sometimes produce much larger shifts in evaluation behavior.
Third, adding more textual descriptions of expert criteria does not always help.
In many cases, injecting explicit criteria and reasoning into model editing actually reduces performance compared to using background information alone.
Finally, some evaluation dimensions remain difficult to align, especially those involving value trade-offs or external knowledge.

Taken together, these findings highlight a fundamental difficulty of expert alignment in subjective tasks.
Disagreement between experts should not always be treated as annotation noise.
Instead, it can reflect genuine diversity in professional judgment.
When evaluation involves experience, value trade-offs, and contextual reasoning, a single stable standard may not exist.
Our study therefore focuses on understanding this diversity rather than eliminating it.
We analyze expert heterogeneity across evaluation dimensions and across types of expert information.
Based on these observations, we discuss the implications for pluralistic evaluation systems that represent multiple expert perspectives rather than enforcing a single evaluation standard.

\section{Related Work}
\label{sec:relatedwork}
\subsection{Human Evaluation, Subjectivity, and Disagreement}

Many NLP and AI evaluation settings rely on human judgments.
This is especially true when quality cannot be measured by exact correctness alone.
Recent work on LLM-as-a-judge shows that language models can correlate well with human preferences in some settings, but also inherit systematic biases and instabilities \cite{Liu2023GEval,Zheng2023}.
These studies are important because they show both the promise and the limits of using LLMs as evaluators.
At the same time, a growing line of work argues that disagreement in human annotation should not always be treated as noise.
CrowdTruth emphasizes that ambiguity and disagreement can reflect meaningful differences in interpretation rather than annotation failure \cite{AroyoWelty2015}.
Davani et al.\ make a similar argument for subjective NLP tasks and show that majority voting can erase important differences between annotators \cite{Davani2022}.
Our work extends this view from general annotators to experts.
In our setting, disagreement is not just a data-cleaning issue.
It is part of the phenomenon we want to understand.

\subsection{Preference Alignment and Personalized LLM Behavior}

Modern alignment methods often assume that there is a target behavior that can be learned from demonstrations or preferences.
A standard example is instruction tuning with human feedback, where models are trained to follow preferred outputs \cite{Ouyang2022}.
More recent work studies how to align models with different users or personas instead of one global target.
PAD performs personalized alignment at decoding time, which avoids retraining a separate model for each preference profile \cite{Chen2024}.
Persona-judge further shows that token-level self-judgment can support personalized alignment without full supervised retraining \cite{Zhang2025}.

These works are closely related to ours because they move from one universal preference to multiple preference profiles.
However, they still usually assume that the target preference is stable enough to be described and optimized.
Our results suggest that this assumption becomes fragile in subjective expert evaluation.
Even when experts provide explicit criteria and reasons, these descriptions may not fully capture the behavior expressed in their scores.

\subsection{Model Editing and Behavioral Control}

Model editing aims to change a model in a targeted way while preserving most of its original knowledge.
ROME shows that factual associations can be localized and edited by modifying a small part of a transformer \cite{Meng2022}.
MEMIT extends this idea to batch editing at scale \cite{Meng2023}.
AlphaEdit adds a null-space constraint to reduce interference and improve sequential editing stability \cite{Fang2024}.

Most editing work focuses on factual knowledge.
Our setting is different.
We use model editing not to update a fact, but to probe whether an expert's evaluation behavior can be injected into a model.
This lets us ask a deeper question: is subjective evaluation behavior sufficiently localized and stable to be edited?
Our results suggest that the answer is often no, or only partially so.
Editing can make small and useful shifts, but it often struggles when the target expert behavior is far from the base model.
This makes the contrast between editing and fine-tuning theoretically meaningful, not just practically useful.

\subsection{Diversity, Values, and Pluralistic Alignment}

A related literature asks whose values or viewpoints language models reflect.
\citet{Santurkar2023} show that language models do not reflect one neutral public opinion, but instead align unevenly with different demographic groups.
\citet{Masoud2025} study cultural alignment and show that model behavior varies across cultural dimensions and prompting conditions.
These findings support a broader view of alignment as a problem of \emph{pluralism}, not just optimization.

Work on generative agents also shows that modeling human behavior depends strongly on the type of information used to represent a person \cite{Park2024}.
This is relevant to our setting because we compare expert background, criteria, and reasoning as different representations of expertise.
Our results show that more explicit information is not always more useful.
In particular, stated criteria and reasoning often help less than expected.
This suggests that expert judgment may depend on tacit, context-dependent knowledge that is difficult to verbalize and difficult to transfer directly.

Overall, our work differs from prior research in one central way.
Rather than assuming that expert evaluation is a fixed target, we ask whether that target is stable enough to align to in the first place.
From this perspective, expert alignment is not only a modeling problem.
It is also a problem of subjective diversity.

\section{Dataset}
\label{sec:dataset}
\subsection{PBIG Expert Evaluation Data}

We build our experiments on the PBIG 2025 shared task: \emph{Product Business Idea Generation from Patents} \cite{Hirota2025}.
In this task, a system generates a business proposal from a patent, and human experts evaluate the proposal along multiple dimensions.
These dimensions cover both technical feasibility and market-oriented judgment.
Following the shared task setting, we use six evaluation dimensions:
\textit{Technical Validity},
\textit{Innovativeness},
\textit{Specificity},
\textit{Need Validity},
\textit{Market Size}, and
\textit{Competitive Advantage} \cite{Hirota2025}.

\begin{table}[t]
\centering
\resizebox{\columnwidth}{!}{
\begin{tabular}{lccc}
\hline
\textbf{Expert Group} & \textbf{\# Experts} & \textbf{Avg. Experience (Years)} & \textbf{\# Test Proposals} \\
\hline
NLP Experts & 3 & 8.0 & 205 \\
IT Experts  & 6 & 5.7 & 233 \\
\hline
Total & 9 & 6.4 & 438 \\
\hline
\end{tabular}
}
\caption{Composition of expert evaluators in the dataset. The experts come from two main backgrounds: NLP research and IT industry practice.}
\label{tab:expert_groups}
\end{table}

This dataset is well suited to our research question for two reasons.
First, it contains judgments from multiple experts rather than a single gold annotator.
Second, several evaluation dimensions are inherently subjective.
For example, \textit{Innovativeness}, \textit{Market Size}, and \textit{Competitive Advantage} often require value trade-offs, outside knowledge, and personal judgment.
This makes the dataset a useful testbed for studying expert alignment under subjective diversity.

\subsection{Questionnaire Data as Structured Expert Descriptions}

Table~\ref{tab:expert_groups} summarizes the participating experts and the number of proposals they evaluated.
Expert scores alone show \emph{what} experts decided, but they do not explain \emph{why} they made those decisions.
To address this gap, we collected a structured questionnaire from each expert.
The questionnaire serves as an additional source of expert information and allows us to study the relation between explicit self-description and actual scoring behavior.

The questionnaire has three parts.

\paragraph{Expert Background:}
We ask each expert about their professional role, years of experience, technical or business expertise, and relevant industry, startup, or investment exposure.
This information captures identity-level characteristics that may shape judgment style.

\paragraph{Stated Evaluation Criteria:}
For each evaluation dimension, we ask experts to describe how they assign scores.
These responses give an explicit verbal account of the standards that experts say they use.

\paragraph{Example-Based Reasoning:}
We present several example proposals and ask experts to provide both scores and short explanations.
These responses offer more concrete traces of how experts justify their decisions in specific cases.

We use these questionnaire responses in two ways.
First, they provide input to our alignment methods, especially model editing.
Second, they support qualitative analysis of expert heterogeneity.
In particular, they let us compare different forms of expert information: background, explicit criteria, and example-based reasoning.
This comparison is important because one of our main findings is that richer verbal explanations do not always lead to better alignment.

\subsection{From Scores to Expert Profiles}

A key feature of our dataset is that it links each expert's scoring records with their own textual descriptions.
This gives us a compact but information-rich expert profile for each participant.
For every expert, we therefore have:
(1) evaluation scores on business proposals,
(2) background information,
(3) self-reported criteria for each dimension, and
(4) reasoning on selected examples.

This design allows us to examine several gaps that are usually hidden in alignment studies.
We can compare:
(a) differences across experts,
(b) differences across evaluation dimensions,
and
(c) differences between what experts score and how they describe their own standards.
These comparisons are central to our framing that expert alignment is difficult not only because of model limitations, but also because expert judgment itself is heterogeneous, partly tacit, and not always easy to verbalize.

\subsection{Expert-Specific Splits}

For each expert, we construct an expert-specific training set and a held-out test set.
The training portion contains only a small number of proposals with expert scores, while the remaining proposals are used for evaluation.
This setup reflects a realistic constraint in expert-driven applications: expert annotations are expensive, and only limited data is usually available for each individual evaluator.

This few-sample setting is especially useful for our study.
It lets us compare methods that make different assumptions about adaptation.
Prompt-based methods test whether a small number of examples is enough to represent an expert's style.
Fine-tuning tests whether broader behavioral adjustment can recover that style.
Model editing tests whether expert behavior can be treated as a more localized form of knowledge.
By keeping the supervision small and expert-specific, the dataset makes these differences visible.

\section{Experiments}
\label{sec:experiment}

Our experiments are designed to answer a central question of this paper:
\emph{how difficult is it to align a language model with individual experts when the evaluation task is subjective?}

To study this question, we compare multiple alignment strategies that differ in how strongly they modify the base model.
We also vary the type of expert information used during alignment.
This design allows us to examine both methodological differences and the role of expert knowledge representations.

\subsection{Base Model}

We use LLaMA-3.1-8B as the base model \cite{Grattafiori2024}.
Given a proposal and an evaluation dimension, the model predicts a discrete score corresponding to the expert's judgment.

\subsection{Alignment Methods}

We compare three families of approaches that represent different levels of adaptation.

\paragraph{Zero-Shot Prompting:}
We prompt the base model with the proposal and the dimension description defined in the PBIG task \cite{Hirota2025}.
The model directly outputs a score without any expert-specific examples.

\paragraph{Few-Shot Prompting:}
We add a small number of expert-labeled examples as in-context demonstrations.
This tests whether a limited set of examples is sufficient to represent an expert's evaluation style.
Our setup follows standard in-context learning practice \cite{Wei2022}.

\paragraph{Fine-Tuning:}
We fine-tune the model on each expert's training set.
We use LoRA to perform parameter-efficient adaptation \cite{Hu2021}.
Compared with prompting, fine-tuning allows larger behavioral changes and can capture more complex expert-specific patterns.

\paragraph{Model Editing:}
We use AlphaEdit to modify the model with a small number of expert examples \cite{Fang2024}.
In this setting, expert evaluation behavior is treated as editable knowledge.
Unlike fine-tuning, model editing aims to make targeted updates while preserving most of the original model.

To better understand what information is useful for editing, we test three editing inputs:

\begin{itemize}
\item \textbf{Background only:} expert identity and experience information.
\item \textbf{Background + Criteria:} background plus the expert's stated evaluation rules.
\item \textbf{Background + Criteria + Reasoning:} the full questionnaire information including example explanations.
\end{itemize}

Our goal is to test whether richer textual descriptions of expert knowledge actually improve alignment.

\subsection{Evaluation Metrics}

Our primary metric is exact-match accuracy.
A prediction is counted as correct if the model outputs the same discrete score as the expert.
Although strict, this metric highlights disagreement clearly.
In later analysis, we also examine distance-based measures that capture near-miss predictions.

\begin{table}[t]
\centering
\small
\setlength{\tabcolsep}{6pt}
\renewcommand{\arraystretch}{1.2}

\resizebox{\columnwidth}{!}{
\begin{tabular}{c|rr|r|r}
 & \multicolumn{2}{c|}{\textbf{Prompting}} & \textbf{Fine-tune} & \textbf{Model Editing} \\
\textbf{Expert} 
 & Zero-shot & Few-shot 
 & LoRA 
 & AlphaEdit (Background) \\
\hline
A & 33.78 & \textbf{41.81} & 34.15 & 33.78 \\
B & 23.19 & \textbf{43.54} & \textbf{43.54} & 31.88 \\
\hline
C & 2.33 & 10.91 & \textbf{29.09} & 23.26 \\
D & 9.38 & 30.41 & \textbf{45.92} & 20.31 \\
\hline
E & 17.54 & 22.78 & 21.51 & \textbf{22.81} \\
F & 15.32 & 22.51 & 27.75 & \textbf{28.23} \\
G & 15.00 & 19.05 & 30.95 & \textbf{31.67} \\
H & 14.44 & 30.25 & 32.77 & \textbf{34.44} \\
I & 14.29 & 22.45 & 33.67 & \textbf{38.57} \\
\end{tabular}
}

\caption{Overall performance (\%) of different alignment strategies across nine experts. }
\label{tab:method_comparison}

\end{table}

\subsection{Main Results}

Table~\ref{tab:method_comparison} compares the overall alignment performance across nine experts.
The table contrasts prompting, fine-tuning, and model editing when only background information is used.
Table~\ref{tab:editing_ablation} then examines how additional expert information affects model editing.

A first pattern is the asymmetry between model editing and fine-tuning.
As shown in Table~\ref{tab:method_comparison}, AlphaEdit sometimes outperforms fine-tuning, but the improvement is typically small.
In contrast, when fine-tuning performs better, the margin can be much larger.
This suggests that model editing primarily provides incremental calibration, while fine-tuning can induce larger behavioral shifts.
In other words, editing appears effective when the expert behavior is close to the base model, whereas fine-tuning is needed when the gap is larger.

A second observation is the strong heterogeneity across experts.
Model performance varies substantially depending on which expert the model attempts to imitate.
Some experts are already relatively close to the base model in the zero-shot setting, and only minor adjustments are needed to approximate their scores.
For other experts, however, the model requires substantial adaptation before its predictions resemble the expert's evaluation behavior.
This variation indicates that expert evaluation styles differ widely in their distance from the base model.

\begin{table}[t]
\centering
\small
\setlength{\tabcolsep}{6pt}
\renewcommand{\arraystretch}{1.2}

\resizebox{\columnwidth}{!}{
\begin{tabular}{c|r|rr}
\textbf{Expert} & \textbf{Background} & \textbf{+Criteria ($\Delta$)} & \textbf{+Reasoning ($\Delta$)} \\
\hline
A & 33.78 & +2.03 & +1.63 \\
B & 31.88 & -2.89 & -5.79 \\
\hline
C & 23.26 & -2.33 & 0.00 \\
D & 20.31 & 0.00 & -4.69 \\
\hline
E & 22.81 & -3.51 & -3.51 \\
F & 28.23 & -0.81 & -3.23 \\
G & 31.67 & -13.34 & -10.00 \\
H & 34.44 & -3.33 & -7.77 \\
I & 38.57 & -4.28 & -12.86 \\
\end{tabular}
}
\caption{Effect of additional expert information in model editing (\%).
Adding explicit criteria or reasoning often decreases accuracy, suggesting that subjective evaluation knowledge is difficult to encode through explicit textual rules.}
\label{tab:editing_ablation}

\end{table}

Finally, Table~\ref{tab:editing_ablation} shows that adding more expert explanations does not necessarily improve editing performance.
Including explicit evaluation criteria or example-based reasoning often reduces accuracy compared with using background information alone.
This result is counter-intuitive if we assume that expert evaluation can be fully captured by explicit rules.
One possible explanation is that subjective evaluation relies on tacit knowledge that is difficult to express in explicit rules.
Experts may articulate part of their reasoning, but their scoring behavior also depends on experience, context, and implicit trade-offs.
In addition, the questionnaire was collected after the original scoring stage, so the written criteria may reflect later rationalization rather than the original decision process.
As a result, injecting these explanations into the model can introduce noise rather than useful guidance.

\section{Analysis: Why Expert Alignment Is Hard}
\label{sec:analysis}

\subsection{Distance to the Base Model}

A central pattern in Table~\ref{tab:method_comparison} is that experts differ greatly in how close they are to the base model.
For some experts, zero-shot prompting already gives moderately strong agreement.
For others, zero-shot performance is very low.
This difference matters because alignment becomes easier when the target expert is already near the model's prior behavior.

This observation helps explain why expert alignment is not a uniform problem.
Some experts require only minor calibration.
In such cases, prompting or model editing can already move the model in the right direction.
Other experts are much farther from the base model.
For them, alignment requires a larger change in evaluation behavior.
This makes adaptation more difficult, especially when the task itself is subjective.

\begin{table}[t]
\centering
\small
\begin{tabular}{lccc}
\hline
\textbf{Method} & \textbf{Exact (0)} & \textbf{Near ($\pm1$)} & \textbf{Far ($\geq2$)} \\
\hline
Zero-shot & 17.9\% & 27.8\% & 54.3\% \\
AlphaEdit & 27.9\% & 30.6\% & 41.6\% \\
\hline
\end{tabular}
\caption{
Distribution of prediction errors measured by absolute score difference.
AlphaEdit increases exact matches and reduces large deviations compared with zero-shot prompting.
}
\label{tab:score_shift_ratio}
\end{table}

Table~\ref{tab:score_shift_ratio} provides a complementary view of this effect.
Compared with zero-shot prompting, AlphaEdit increases the proportion of exact matches from 17.9\% to 27.9\% and slightly increases near matches from 27.8\% to 30.6\%.
At the same time, it reduces the proportion of large deviations ($\geq 2$ score points) from 54.3\% to 41.6\%.
The main gain is therefore not only exact matching.
Model editing also improves calibration by shifting more predictions closer to expert judgments.

This pattern suggests a simple but important interpretation:
expert alignment difficulty depends strongly on how far an expert's evaluation behavior deviates from the base model's prior.
When the gap is small, local calibration may be enough.
When the gap is large, alignment becomes much harder.

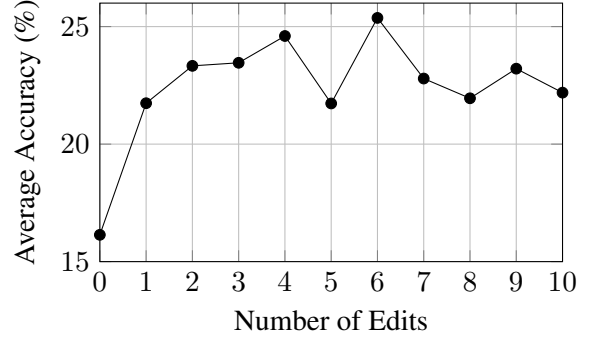
\begin{figure}[t]
\centering
\begin{tikzpicture}
\begin{axis}[
width=\linewidth,
height=5cm,
xlabel={Number of Edits},
ylabel={Average Accuracy (\%)},
xmin=0, xmax=10,
ymin=15, ymax=26,
xtick={0,1,2,3,4,5,6,7,8,9,10},
grid=both,
mark size=2pt,
]

\addplot[
color=black,
mark=*,
]
coordinates {
(0,16.14)
(1,21.74)
(2,23.33)
(3,23.46)
(4,24.60)
(5,21.73)
(6,25.37)
(7,22.79)
(8,21.95)
(9,23.21)
(10,22.19)
};

\end{axis}
\end{tikzpicture}

\caption{
Average alignment accuracy as a function of the cumulative number of edits using AlphaEdit with background information.
}
\label{fig:editing_steps}

\end{figure}

\subsection{Temporal Instability of Expert Judgment}

A further source of difficulty is the temporal instability of expert judgments.
To examine this effect, we asked experts to re-evaluate a set of proposals approximately three months after the initial evaluation stage.

The scores are not always identical across the two evaluation rounds.
Among the re-evaluated cases, only 37.5\% of the scores remained unchanged.
The remaining cases shifted by one score point, with 25.0\% decreasing and 37.5\% increasing.
This suggests that even the same expert may revise their judgment over time.

This observation highlights a fundamental challenge for expert alignment.
Most alignment approaches assume that the target behavior is stable and well defined.
However, when expert judgments themselves evolve over time, the alignment target becomes moving rather than fixed.

This temporal drift also helps explain why explicit criteria collected in the questionnaire do not always improve alignment.
If the criteria reflect updated interpretations or later rationalizations, they may no longer correspond exactly to the decision process used during the earlier evaluation stage.
As a result, the model may be trained to match a description of expert behavior that differs slightly from the behavior reflected in the dataset.

More broadly, this finding suggests that subjective evaluation may not have a single static standard.
Instead, expert judgments can shift as evaluators accumulate experience or reconsider their criteria.
For alignment systems, this implies that learning a fixed expert rule may be inherently difficult.

\begin{figure}[t]
\centering
\begin{tikzpicture}
\begin{axis}[
width=\linewidth,
height=5.2cm,
xlabel={Single Edit Sample (sorted by mean gain)},
ylabel={Mean Accuracy Gain ($\Delta$)},
xmin=0.5, xmax=10.5,
ymin=-2, ymax=18,
xtick={1,...,10},
xticklabels={S1,S2,S3,S4,S5,S6,S7,S8,S9,S10},
ytick={0,4,8,12,16},
ymajorgrids=true,
xmajorgrids=false,
grid style={dashed,gray!25},
tick label style={font=\small},
label style={font=\small},
axis line style={black},
tick style={black},
]

\addplot[
only marks,
color=black,
mark=*,
mark size=2.4pt,
error bars/.cd,
y dir=both,
y explicit,
error bar style={line width=0.7pt},
]
coordinates {
(1,2.54) +- (0,4.37)
(2,3.93) +- (0,4.61)
(3,4.93) +- (0,5.51)
(4,5.33) +- (0,6.50)
(5,5.60) +- (0,6.67)
(6,6.16) +- (0,4.52)
(7,6.53) +- (0,4.95)
(8,6.58) +- (0,5.35)
(9,7.31) +- (0,6.41)
(10,7.97) +- (0,7.93)
};

\end{axis}
\end{tikzpicture}

\caption{
Mean accuracy gain from a single edit example relative to baseline, aggregated across experts.
Edit samples are sorted by mean gain.
Error bars denote standard deviation across experts.
}
\label{fig:single_edit_stats}
\end{figure}
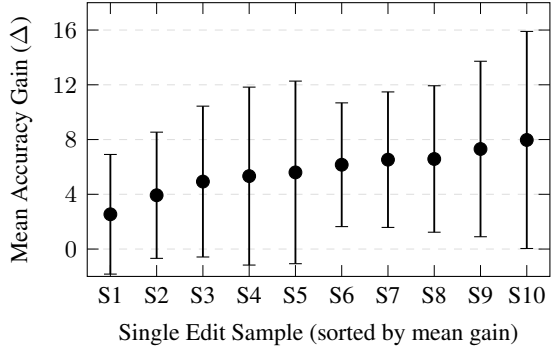

\subsection{Sample Efficiency and Sample Selection}

Another notable pattern is that editing is highly sensitive to both the number and the identity of examples.
Figure~\ref{fig:editing_steps} shows performance as a function of the cumulative number of edits.
The first edit already produces a substantial improvement, increasing average accuracy from 16.14\% to 21.74\%.
Additional edits yield smaller and less stable gains.
Although the best average performance appears after six edits (25.37\%), later edits do not consistently improve results and instead fluctuate around a similar range.
This suggests that editing can efficiently shift model behavior with only a small number of examples, but it is less effective for accumulating larger amounts of subjective evaluation knowledge through repeated local updates.

Figure~\ref{fig:single_edit_stats} further shows that sample selection matters greatly.
Across the ten single-edit examples, the mean gain relative to baseline ranges from 2.54 to 7.97 accuracy points.
At the same time, the large error bars indicate substantial variation across experts even for the same edit sample.
In other words, some examples provide broadly useful calibration signals, whereas others help only certain experts.
This indicates that the alignment signal is sparse and unevenly distributed, and that a small number of high-leverage cases can dominate the effect of editing.

This sensitivity is another reason why expert alignment is hard.
When the target behavior is subjective, the model does not simply need more data.
It needs the right data.
A few representative cases may be enough to calibrate the model, but poorly chosen examples may fail to capture the expert's actual decision boundary.

\begin{table}[t]
\centering
\small
\begin{tabular}{lrr}
\hline
\textbf{Dimension} & \textbf{Mean Acc.} & \textbf{Std} \\
\hline
Technical Validity & 36.37 & 14.95 \\
Market Size & 34.47 & 18.71 \\
Specificity & 33.41 & 22.52 \\
\hline
Competitive Advantage & 9.86 & 7.29 \\
Need Validity & 6.13 & 9.89 \\
Innovativeness & 4.10 & 4.36 \\
\hline
\end{tabular}
\caption{
Alignment accuracy aggregated across experts and editing steps for each evaluation dimension.
}
\label{tab:dimension_difficulty}
\end{table}

\begin{table}[t]
\centering
\small
\begin{tabular}{lrrr}
\hline
\textbf{Dimension} & \textbf{Baseline} & \textbf{Best} & \textbf{Gain} \\
\hline
Technical Validity & 18.17 & 44.53 & +26.36 \\
Market Size & 20.24 & 41.19 & +20.95 \\
Specificity & 29.71 & 37.60 & +7.88 \\
\hline
Innovativeness & 1.19 & 6.64 & +5.45 \\
Need Validity & 4.12 & 8.89 & +4.77 \\
Competitive Advantage & 11.73 & 12.92 & +1.19 \\
\hline
\end{tabular}
\caption{
Editing improvement across evaluation dimensions.
Baseline denotes accuracy before editing, Best denotes the highest average accuracy across editing steps,
and Gain denotes the improvement relative to baseline.
}
\label{tab:dimension_edit_gain}
\end{table}

\subsection{Dimension-Level Difficulty}

Alignment difficulty also differs substantially across evaluation dimensions.
Table~\ref{tab:dimension_difficulty} summarizes accuracy statistics aggregated across experts and editing steps.
The results reveal a clear separation between dimensions that can be evaluated primarily from the proposal text and those that require broader contextual reasoning.

Dimensions such as \textit{Technical Validity}, \textit{Market Size}, and \textit{Specificity} achieve substantially higher accuracy overall, with mean accuracies of 36.37\%, 34.47\%, and 33.41\%, respectively.
These dimensions often rely on evidence that can be directly inferred from the proposal content, such as technical feasibility or the clarity of the described idea.
In contrast, \textit{Competitive Advantage}, \textit{Need Validity}, and \textit{Innovativeness} remain much more difficult to align, with mean accuracies below 10\%.
Evaluating these dimensions typically requires external knowledge, market reasoning, or implicit value judgments that are not fully captured by the proposal text alone.

Table~\ref{tab:dimension_edit_gain} further shows that the impact of model editing also varies substantially across dimensions.
Editing produces large improvements for \textit{Technical Validity} and \textit{Market Size}, increasing accuracy by 26.36 and 20.95 percentage points relative to baseline.
A smaller but still noticeable improvement appears for \textit{Specificity} (+7.88).
By contrast, improvements remain limited for \textit{Competitive Advantage}, \textit{Need Validity}, and \textit{Innovativeness}, where gains remain below 6 percentage points.

Taken together, these results suggest a subjectivity gradient in expert evaluation.
When a dimension can be judged primarily from textual evidence in the proposal, alignment is both easier and more responsive to editing.
When evaluation depends on broader contextual knowledge or subjective value trade-offs, alignment becomes substantially harder and editing provides only limited benefits.
This pattern indicates that the difficulty of expert alignment arises not only from disagreement between experts, but also from the intrinsic nature of the evaluation dimension itself.

\section{Discussion}
\label{sec:discussion}

Our results support a broader view of expert alignment in subjective tasks.
The main challenge is not only to build a stronger adaptation method.
It is to understand what kind of target expert judgment actually is.

\subsection{Subjective Evaluation as Diverse Judgment}

A common assumption in alignment research is that disagreement reflects noise around an underlying target.
Our findings suggest a different view for subjective evaluation.
Different experts can apply different but still reasonable standards.
Their disagreement is therefore not always a problem to be removed.
In many cases, it reflects legitimate diversity in professional judgment.

This interpretation is consistent with the setting of our task.
Business proposal evaluation requires multiple forms of judgment, including technical plausibility, novelty, need assessment, and market reasoning.
These are not purely objective labels.
They depend on perspective, background, and implicit priorities.
As a result, a single stable expert standard may not always exist.

\subsection{When Is Editing Enough?}
Model editing can be attractive when the target expert is already close to the base model and the goal is small-scale calibration.
In such cases, AlphaEdit can achieve performance close to fine-tuning, and sometimes even slightly better, while modifying the model more locally.

At the same time, our findings also show the limits of this approach.
Editing does not appear to fully reconstruct expert behavior when the gap between the expert and the base model is large.
For those cases, fine-tuning remains more effective.
A useful practical view is therefore to treat editing as a low-cost calibration method and fine-tuning as a higher-capacity adaptation method.

\subsection{From Single-Expert Alignment to Pluralistic Evaluation}

The broader implication of our study is that subjective evaluation may be better framed as a pluralistic problem.
Instead of forcing a model to imitate one ``correct'' expert, it may be more useful to model a distribution of expert perspectives.
Such a system could represent disagreement explicitly, indicate uncertainty, and surface multiple plausible judgments rather than only one score.

This perspective also changes how we interpret alignment failure.
If experts themselves differ in stable and meaningful ways, then imperfect alignment is not always evidence of model weakness alone.
It can also reflect the structure of the task.
For subjective evaluation, the goal may not be perfect agreement with one evaluator, but faithful representation of diverse expert views.

\section{Conclusion}
\label{sec:conclusion}

We studied expert alignment as a way to better understand subjective evaluation. 
We compared prompting, fine-tuning, and model editing across nine experts.
Our analysis reveals four key findings.
First, alignment difficulty varies substantially across experts, indicating that expert evaluation styles differ widely in their distance from the base model. 
Second, explicit criteria and reasoning do not consistently improve alignment, suggesting that expert judgment cannot be fully captured by verbalized rules. 
Third, editing is highly sensitive to both the number and the identity of examples: a small number of edits can provide useful calibration, but improvements are unstable and depend strongly on sample selection. 
Fourth, alignment difficulty differs across evaluation dimensions, with dimensions grounded in proposal content being easier to align than those requiring external knowledge or value-based judgment.

These results suggest that expert alignment is difficult not only because of model limitations, but also because subjective evaluation itself is heterogeneous, partly tacit, and dimension-dependent. 
Rather than assuming a single stable expert standard, future alignment systems may need to account for diverse expert perspectives and evolving evaluation criteria.

\section*{Limitations}

First, our experiments involve a relatively small number of experts in a single task setting.
While this setup allows us to examine subjective evaluation in depth, the findings may not directly generalize to other domains, larger expert populations, or tasks with different types of expertise.

Second, we evaluate alignment using a single base model.
Different model families or larger models may exhibit different alignment behaviors, particularly in how they respond to editing or fine-tuning.
Our results should therefore be interpreted as evidence from one representative model setting rather than a universal claim about all language models.

Finally, we only evaluate one model editing method, AlphaEdit.
Although many editing techniques have been proposed, we adopt AlphaEdit because it represents a strong and stable editing framework in recent work and is designed to reduce unintended side effects on unrelated model knowledge.
Our goal is not to benchmark editing algorithms, but to use a representative editing method to study expert alignment behavior in subjective evaluation.
Future work could extend this analysis by comparing multiple editing approaches.

\bibliography{custom}

@article{Chen2024,
  author    = {Chen, Ruizhe and Zhang, Xiaotian and Luo, Meng and Chai, Wenhao},
  title     = {Pad: Personalized alignment of llms at decoding-time},
  journal   = {arXiv preprint arXiv:2410.04070},
  year      = {2024}
}

@article{Hu2021,
  author    = {Hu, Edward J. and Shen, Yelong and Wallis, Phil and Allen{-}Zhu, Zeyuan and Li, Yuanzhi and Wang, Shean and Wang, Lu and Chen, Weizhu},
  title     = {LoRA: Low-rank Adaptation of Large Language Models},
  journal   = {arXiv preprint arXiv:2106.09685},
  year      = {2021}
}

@article{Fang2024,
  author    = {Fang, Junfengand Jiang, Houcheng and Wang, Kun and Ma, Yunshan and Jie, Shi and Wang, Xiangnan He and Chua, Tat-seng},
  title={AlphaEdit: Null-Space Constrained Knowledge Editing for Language Models},
  journal={The Thirteenth International Conference on Learning Representations},
  year      = {2025}
}

@article{Grattafiori2024,
  author    = {Grattafiori, Anthony and Dubey, Apoorv and Jauhri, Arun and Pandey, Atul and Kadian, Aditya and Al{-}Dahle, Ahmad and Letman, Andrew and Mathur, Anuj and Schelten, Anias and Vaughan, Andrew and Yang, Anu and Fan, Angela and Goyal, Aniruddha and Hartshorn, Anna and Yang, Artemis and Mitra, Arnav and Sravankumar, Ashok and Korenev, Artem and Hinsvark, Asmund and Rao, Atul and Zhang, Ayan and Rodriguez, Alejandro and Gregerson, Alexander and Spataru, Alex and Roziere, Baptiste and Biron, Benjamin and Tang, Bo and Chern, Boris and Caucheteux, Clara and Nayak, Chandan and Bi, Cangxiong and Marra, Chris and McConnell, Charles and Keller, Claude and Touret, Clement and Wu, Chen and Wong, Chau{-}Wai and Ferrer, Claudia Castellon and Nikolaidis, Christos and Allonsius, Daniel and Song, Danqi and Pintz, Daniel and Livshits, David and Wyatt, David and Esiobu, Destiny and Choudhary, Dipanjan and Mahajan, Divyansh and Garcia{-}Olano, Diego and Perino, Dino and Hupkes, Dieuwke and Lakomkin, Egor and AlBadawy, Eman and Lobanova, Evgeniia and Dinan, Emily and Smith, Emily M. and Radenovic, Filip and Guzm{\'a}n, Francisco and Zhang, Fuzhao and Synnaeve, Guillaume and Lee, Geunsik and Anderson, Glenn L. and Thattai, Govindan and Nail, Grant and Mialon, Gr{\'e}goire and Pang, Guanming and Cucurell, Guillem and Nguyen, Hieu and Korevaar, Henrike and Xu, Hu and Touvron, Hugo and Zarov, Ibraheem and Ibarra, Ignacio A. and Kloumann, Ilse and Misra, Ishan and Evtimov, Ivan and Zhang, Jiageng and Copet, Julien and Lee, Junbum and Geffert, Julius and Vranes, Justin and Park, Junsu and Mahadeokar, Jyothi and Shah, Karan and van der Linde, Kasper and Billock, Kevin and Hong, Kiseok and Lee, Kyeongmin and Fu, Liwei and Chi, Meng and Huang, Meng and Liu, Mengdi and Wang, Menghui and Yu, Mingxing and Bitton, Mor and Spisak, Nicholas and Park, Noel and Rocca, Nicola and Johnstun, Nikolai and Saxe, Neal and Jia, Nicholas and ... (460 additional authors)},
  title     = {The Llama 3 Herd of Models},
  journal   = {arXiv preprint arXiv:2407.21783},
  year      = {2024}
}

@inproceedings{Hirota2025,
  author    = {Hirota, Wataru and Chen, Chung-chi and Ohkuma, Tomoko and Taniguchi, Tomoki and Ishigaki, Tatsuya},
  title     = {Overview of PBIG Shared Task at AgentScen 2025: Product Business Idea Generation from Patents},
  booktitle = {Proceedings of the 2nd Workshop on Agent AI for Scenario Planning},
  year      = {2025},
  pages     = {35-42}
}

@inproceedings{Masoud2025,
  author    = {Masoud, Reem I. and Liu, Ziquan and Ferianc, Martin and Treleaven, Philip C. and Rodrigues, Miguel},
  title     = {Cultural Alignment in Large Language Models: An Explanatory Analysis Based on Hofstede’s Cultural Dimensions},
  booktitle = {Proceedings of the 31st International Conference on Computational Linguistics},
  year      = {2025},
  pages     = {8474-8503}
}

@inproceedings{Meng2022,
  author    = {Meng, Kevin and Bau, David and Andonianm, Alex and Belinkov, Yonatan },
  title     = {Locating and editing factual associations in GPT},
  booktitle = {Proceedings of the 36th International Conference on Neural Information Processing Systems (NIPS '22)},
  year      = {2022},
  pages     = {17359-17372}
}

@article{Meng2023,
  author    = {Meng, Kevin and Sharma, Arnab Sen and Andonianm, Alex and Belinkov, Yonatan and Bau, David},
  title     = {Mass-Editing Memory in a Transformer},
  journal   = {arXiv preprint arXiv:2210.07229},
  year      = {2023}
}

@article{Park2024,
  author    = {Park, Joon Sung and Zou, Carolyn Q. and Shaw, Aaron and Hill, Benjamin Mako and Cai, Carrie and Morris, Meredith Ringel and Willer, Robb and Liang, Percy and Bernstein, Michael S.},
  title     = {Generative Agent Simulations of 1,000 People},
  journal   = {arXiv preprint arXiv:2411.10109},
  year      = {2024}
}

@inproceedings{Zhang2025,
  author    = {Zhang, Xiaotian and Chen, Ruizhe and Feng, Yang and Liu, Zuozhu},
  title     = {Persona-judge: Personalized Alignment of Large Language Models via Token-level Self-judgment},
  booktitle = {Findings of the Association for Computational Linguistics: ACL 2025},
  year      = {2025},
  pages     = {5037-5049}
}

@article{Wei2022,
  author    = {Wei, Jason and Bosma, Maarten and Zhao, Vincent Y. and Guu, Kelvin and Yu, Adams Wei and Lester, Brian and Du, Nan and Dai, Andrew M. and Le, Quoc V.},
  title     = {Finetuned language models are zero-shot learners},
  journal   = {arXiv preprint arXiv:2109.01652},
  year      = {2022}
}

@inproceedings{Ouyang2022,
  title     = {Training Language Models to Follow Instructions with Human Feedback},
  author    = {Ouyang, Long and Wu, Jeffrey and Jiang, Xu and Almeida, Diogo and Wainwright, Carroll and Mishkin, Pamela and Zhang, Chong and Agarwal, Sandhini and Slama, Katarina and Ray, Alex and Schulman, John and Hilton, Jacob and Kelton, Fraser and Miller, Luke and Simens, Maddie and Askell, Amanda and Welinder, Peter and Christiano, Paul F. and Leike, Jan and Lowe, Ryan},
  booktitle = {Advances in Neural Information Processing Systems},
  year      = {2022}
}

@inproceedings{Liu2023GEval,
  title     = {G-Eval: NLG Evaluation using GPT-4 with Better Human Alignment},
  author    = {Liu, Yang and Iter, Dan and Xu, Yichong and Wang, Shuohang and Xu, Ruochen and Zhu, Chenguang},
  booktitle = {Proceedings of the 2023 Conference on Empirical Methods in Natural Language Processing},
  year      = {2023},
  pages     = {2511--2522}
}

@article{Zheng2023,
  title   = {Judging LLM-as-a-Judge with MT-Bench and Chatbot Arena},
  author  = {Zheng, Lianmin and Chiang, Wei-Lin and Sheng, Ying and Zhuang, Siyuan and Wu, Zhanghao and Zhuang, Yonghao and Lin, Zi and Li, Zhuohan and Li, Dacheng and Xing, Eric P. and Zhang, Hao and Gonzalez, Joseph E. and Stoica, Ion},
  journal = {arXiv preprint arXiv:2306.05685},
  year    = {2023}
}

@article{AroyoWelty2015,
  title   = {Truth Is a Lie: Crowd Truth and the Seven Myths of Human Annotation},
  author  = {Aroyo, Lora and Welty, Chris},
  journal = {AI Magazine},
  year    = {2015},
  volume  = {36},
  number  = {1},
  pages   = {15--24}
}

@inproceedings{Santurkar2023,
  author    = {Shibani Santurkar and Esin Durmus and Faisal Ladhak and Cinoo Lee and Percy Liang and Tatsunori Hashimoto},
  title     = {Whose Opinions Do Language Models Reflect?},
  booktitle = {Proceedings of the 40th International Conference on Machine Learning},
  series    = {Proceedings of Machine Learning Research},
  volume    = {202},
  pages     = {29971--30004},
  year      = {2023},
  publisher = {PMLR}
}

@article{Davani2022,
  title   = {Dealing with Disagreements: Looking Beyond the Majority Vote in Subjective Annotations},
  author  = {Davani, Aida Mostafazadeh and Díaz, Mark and Prabhakaran, Vinodkumar},
  journal = {Transactions of the Association for Computational Linguistics},
  year    = {2022},
  volume  = {10},
  pages   = {92--110}
}

\end{document}